\documentclass[runningheads]{llncs}
\usepackage{comment}
\usepackage{graphicx}
\usepackage{subfigure}
\usepackage{color}
\usepackage{amsfonts,amssymb}
\usepackage[hidelinks]{hyperref} 
\usepackage{multirow}

\begin{document}
\newcommand{\myred}[1]{\textcolor{red}{#1}}
\newcommand{\ask}[1]{\textcolor{blue}{#1}}

\newcommand{\sname}{\emph{Graph-EKLN}}
\newcommand{\fname}{\emph{Graph-based Exercise- and Knowledge-Aware Learning Network}}
\title{Graph-based Exercise- and Knowledge-Aware Learning Network for Student Performance Prediction}

\def\YOFOSubNumber{69}  

\titlerunning{Graph-based Exercise- and Knowledge-Aware Learning Network}
%
\author{Mengfan Liu\inst{1,2} \and
Pengyang Shao\inst{1,2} \and
Kun Zhang\inst{1,2,}\thanks{Kun Zhang is the corresponding author.}
}

\authorrunning{M. Liu et al.}
%
\institute{Key Laboratory of Knowledge Engineering with Big Data, Hefei University of Technology, China\and 
School of Computer Science and Information Engineering, Hefei University of Technology, China\\
\email{\{liumengff,shaopymark,zhang1028kun\}@gmail.com}}
\maketitle              

\begin{abstract}
Predicting student performance is a fundamental task in Intelligent Tutoring Systems~(ITSs), by which we can learn about students' knowledge level and provide personalized teaching
strategies for them. 
Researchers have made plenty of efforts on this task. 
They either leverage educational psychology methods to predict students' scores according to the learned knowledge proficiency, or make full use of Collaborative Filtering (CF) models to represent latent factors of students and exercises. 
However, most of these methods either neglect the exercise-specific characteristics (e.g., exercise materials), or cannot fully explore the high-order interactions between students, exercises, as well as knowledge concepts. 
To this end, we propose a \fname~for accurate student score prediction. 
Specifically, we learn students' mastery of exercises and knowledge concepts respectively to model the two-fold effects of exercises and knowledge concepts. Then, to model the high-order interactions, we apply graph convolution techniques in the prediction process. Extensive experiments on two real-world datasets prove the effectiveness of our proposed \sname.

\keywords{Education Data Mining  \and Intelligent Tutoring System \and Collaborative Filtering \and Graph Neural Network}
\end{abstract}
\section{Introduction}
\label{intro}
Intelligent Tutoring Systems~(ITSs) aim at providing personalized guidance for students~\cite{engaging,itsbook}, which can be treated as an important supplementary for traditional offline teaching mode. 
It has attracted enormous attention from both industry and academics~\cite{addressing,edm}. 

Usually, researchers consider the issue from the educational psychology perspective and propose cognitive diagnosis models to discover students' knowledge proficiency~\cite{engaging}. 
Among them, the Deterministic Inputs, Noisy "And" Gate (DINA) model is a representative method which uses multi-dimensional factors to represent students' knowledge states on specific knowledge concepts~\cite{dina}. However, they ignore the influence of other exercise-specific characteristics. 
knowledge proficiency is not the only factor that affects students' final scores. For example, exercise materials can also influence exercises' difficulty~\cite{ekt}.

Moreover, motivated by the observation that students and exercises are collaboratively correlated, 
researchers borrow success of Matrix Factorization~(MF) techniques in recommender systems to model the interactions between students and exercises~\cite{rspsp,multi}. 
For example, In~\cite{rspsp}, the authors applied MF to learn the latent embeddings of students and exercises and predicted the scores based on the inner products of them. Although MF based models achieve great success in ITSs, they still have some weaknesses. 
First of all, MF based methods are still inadequate in utilizing knowledge concept information, which is very important for student performance prediction.
Second, MF based methods cannot deal with the high-order collaborative information between student and exercises. 

\begin{figure}
 \centering
 \scalebox{0.95}{
    \includegraphics[width=\textwidth]{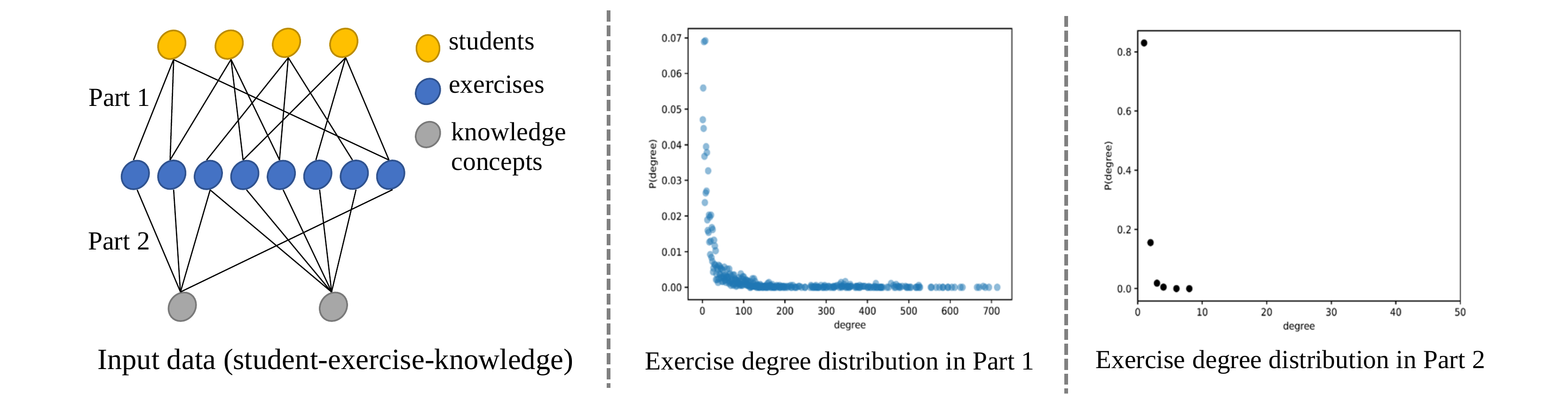}
}
\caption{Data structure of ITSs}
\label{fig:hetero}
\end{figure}

Since students and exercises naturally form a bipartite graph structure, 
it is natural to apply Graph Convolutional Network~(GCN) to model the high-order collaborative information in the student-exercise-knowledge graph. 
However, different from scale-free networks, distribution of exercise-knowledge data does not satisfy the power law distribution~(i.e., shown in the middle part of Fig.~\ref{fig:hetero}). 
More specifically, the degree distribution of exercise is uniform, i.e., most exercises are related to around 1 to 2 knowledge concepts, as shown in the right part of Fig.~\ref{fig:hetero}. 
Furthermore, the number of knowledge concepts is relatively small. Therefore, knowledge concept nodes would link to many exercise nodes, leading to over smoothing when propagating information through dense links. 
To this end, in this paper, we propose a novel \fname~(\sname), which takes the both influences of exercises and knowledge concepts into consideration for student performance prediction. 
For the effect of exercises, we apply GCN with link-specific aggregation functions~\cite{rgcn} onto the student-exercise bipartite graph to explore the high-order collaborative information.
For the effect of knowledge concepts, we replace exercises with their related knowledge concepts, and predict students' performance scores on knowledge concepts by MF. 
Along this line, the high-order graph structure information and knowledge concept information can be fully explored for the final student performance prediction. 
\section{Related Work}
\subsection{Educational Psychology}
Educational psychology models are mainly discussed from two sides: cognitive diagnosis models and knowledge tracing models~\cite{ekt,rkt}. Cognitive diagnosis models, assuming students' knowledge states are static throughout their practice, aim to discover students' proficiency to predict their future performance~\cite{fuzzy,itsbook,wang2020neural,Mt-mcd,dina,dirt}. Item Response Theory (IRT)~\cite{irt,itsbook} was a typical and straightforward cognitive diagnosis model which used a one-dimensional continuous variable $\theta $ to indicate each student's knowledge state and used $\beta$ to indicate each exercise's difficulty. In this way, $(\theta-\beta)$ was proportional to the predicted probability of the question being answered correctly. 
Another typical model was DINA~\cite{dina}. DINA was a multi-dimensional discrete model to represent each student with a binary latent vector. We can know whether the student has mastered related knowledge concepts from students' knowledge states ('1' indicates the student has mastered the target knowledge concepts and vice versa). 
Recently, deep neural networks have been used for cognitive diagnosis. For example, Cheng et al. leveraged deep learning to enhance the process of diagnosing parameters~\cite{dirt}. Wang et al. proposed to incorporate neural networks to learn interaction functions between students and exercises~\cite{wang2020neural}. Knowledge tracing models aims to track the changes of students' knowledge states during practice~\cite{bkt,ekt,learningorforgetting,convolutionalkt}. 
Researchers proposed a first-order Markov process model, in which knowledge states will change with transition probabilities after a learning opportunity~\cite{bkt}. 
Piech et al. introduced a recurrent neural network to describe the change of knowledge states~\cite{dkt}. Liu et al. explored the text content of exercises by integrating a bidirectional LSTM model~\cite{ekt}.
\subsection{Collaborative Filtering in Recommender Systems}
Recommender systems have been widely utilized to help users find their potential interests in many areas~\cite{an2019neural,chen2021set2setrank,wu2021survey}.
Classical models utilize MF techniques to learn user and item embeddings~\cite{mf}. 
Motivated by the observation that users and items naturally form a bipartite graph, researchers proposed to utilize Graph Convolution Networks (GCNs) to model high-order collaborative signals in recommender systems~\cite{lrgccf,rgcn,wu2020diffnet++,diffnet,wu2020learning,wu2020joint}.  
E.g., Wang et al. used the graph convolution technique to encode collaborative signals in the propagation process~\cite{ngcf}. Wu et al. modeled social diffusion process by propagating embeddings in the social network~\cite{diffnet}. 
Chen et al. enhanced graph based recommendation  by empirically removing non-linearities and proposed a residual network based structure~\cite{lrgccf}.  
\subsection{Collaborative Filtering in ITS} 
Motivated by the observation that students and exercises are collaboratively correlated, researchers mapped educational data to user-item-rating triple data in recommender systems, then applied MF for predicting student performance~\cite{rspsp}. 
To improve prediction results, Thai et al. proposed MRMF to explore the multiple relationships between students, exercises, and knowledge concepts by MF techniques~\cite{multi}. Similarly, CRMF integrated the course relationships to update representations of exercises~\cite{crmf}. Moreover, researchers were inspired by social recommendation systems and used the SocialMF technique to improve the prediction accuracy~\cite{social}. 
Furthermore, Nakagawa et al. proposed GKT that viewed knowledge concepts and their dependencies as nodes and links in a graph, so that students' knowledge states on the answered concepts and their related concepts can be both updated over time~\cite{gkt}. 
Note that, students and exercises naturally form an interaction graph in ITSs. 
Considering that GCN can enhance recommendation performance in the user-item bipartite graph, we aim to propose a model that applies GCN onto the student-exercise bipartite graph in ITSs. 
\section{The Proposed Model}
\subsection{Problem Formulation}
Suppose there are $M$ students, $N$ exercises, and $O$ knowledge concepts. Interactions between students and exercises are represented with matrix $\mathbf{R}=\{r_{sp}\}_{M\times N}$, where $r_{sp}$ represents the performance score that student $s$ has on exercise $p$. 
In most cases, the observed part of $\mathbf{R}$ consists of 0 and 1, where $r_{sp}=1$ if student $s$'s answer to exercise $p$  is correct and $r_{sp}=0$ otherwise. 
As for relations between exercises and knowledge concepts, educational experts manually label each exercise with several knowledge concepts. We use matrix $\mathbf{Q}=\{q_{pk}\}_{N\times O}$  to denote the relations, where $q_{pk} = 1$ if exercise $p$ is related to knowledge concept $k$ and $q_{pk} = 0$ if there are no relations between them. 
Given observed interactions $\mathbf{R}$ and relations between exercises and knowledge concepts $\mathbf{Q}$, we aim to predict unobserved $\hat{r}_{sp}$, namely student $s$'s score on non-interactive exercise $p$.  
\subsection{Overall Structure}
\label{overall_s}
\begin{figure}
\includegraphics[width=\textwidth]{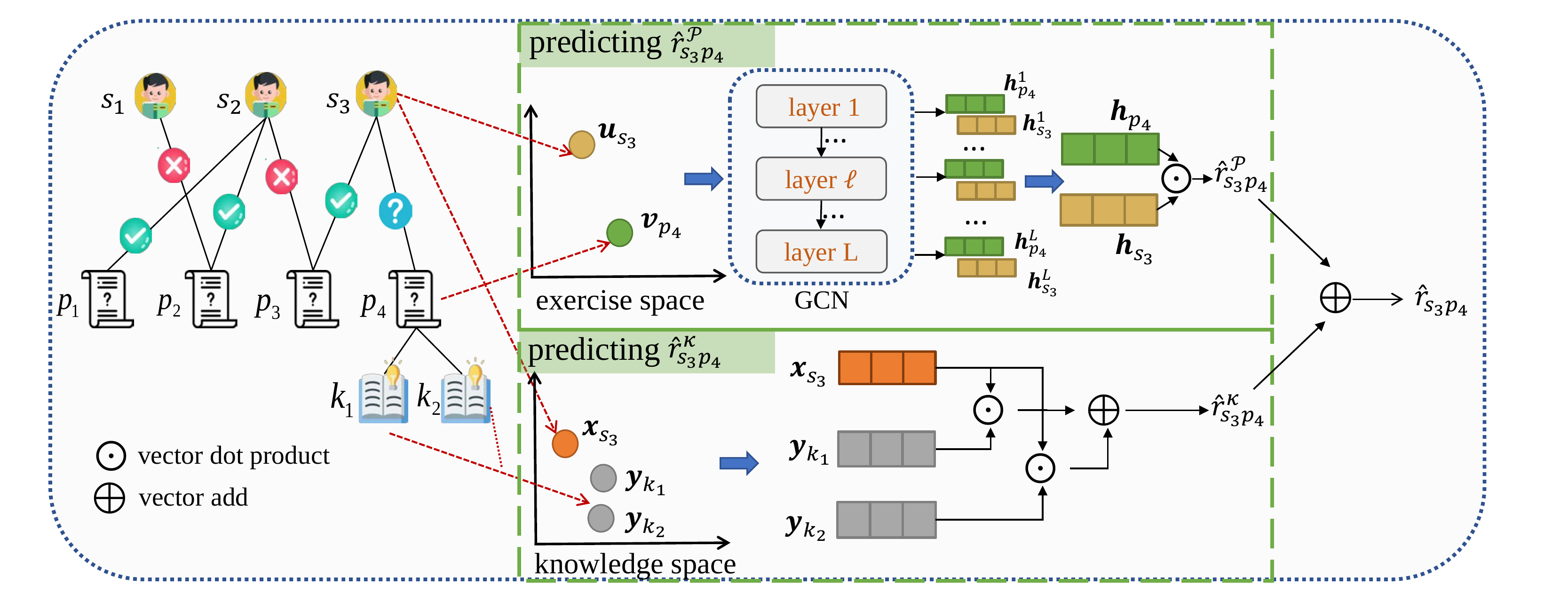}
\caption{The overall structure of our proposed \sname. 
 } \label{fig1}
\end{figure}

The overall structure of \fname~(\sname) is shown in Fig. \ref{fig1}. In the left part of Fig. \ref{fig1}., links between students and exercises are established according to matrix $R$ and links between exercises and knowledge concepts are established according to matrix $Q$. 
There two challenges in our task: how to handle with the different links (correct  answer/wrong answer) between students and  exercises and how to utilize knowledge concept information in MF based models.

To address these two challenges, we divide the task into two sub-tasks, as shown in the middle part of Fig. \ref{fig1}. 
The first sub-task is to predict student $s$'s proficiency on exercise $p$ itself  $\hat{r}_{sp}^\mathcal{P}$.
Note that, $\mathcal{P}$ denotes the predicted score $\hat{r}_{sp}^\mathcal{P}$ is in the exercise space.  
The second sub-task is to predict a student's proficiency on an exercise's related knowledge concepts $\hat{r}_{sp}^\mathcal{K}$. 
Note that, $\mathcal{K}$ denotes the predicted score $\hat{r}_{sp}^\mathcal{K}$ is in the knowledge space.  The following two subsections describe the two sub-tasks respectively.
\subsection{Modeling High-order Collaborative Information}
\label{sec3.3}
Student embeddings and exercise embeddings in the exercise space are represented with $ \mathbf{U} = [\mathbf{u}_1,...,\mathbf{u}_s,..., \mathbf{u}_M] \in \mathbb{R}^{M\times D}$, $ \mathbf{V} = [\mathbf{v}_1,...,\mathbf{v}_p,...,\mathbf{v}_N] \in \mathbb{R}^{N\times D}$ respectively. $D$ denotes the embedding size and $\mathbf{u}_s ,\mathbf{v}_p$ represent the initial embeddings of student $s$ and exercise $p$. 
To address the different-link problem mentioned in subsection \ref{overall_s},  we follow R-GCN~\cite{rgcn}, which is proposed to learn representations of multi-relational graph. 

\begin{figure}
\centering
\scalebox{0.8}{
\includegraphics[width=\textwidth]{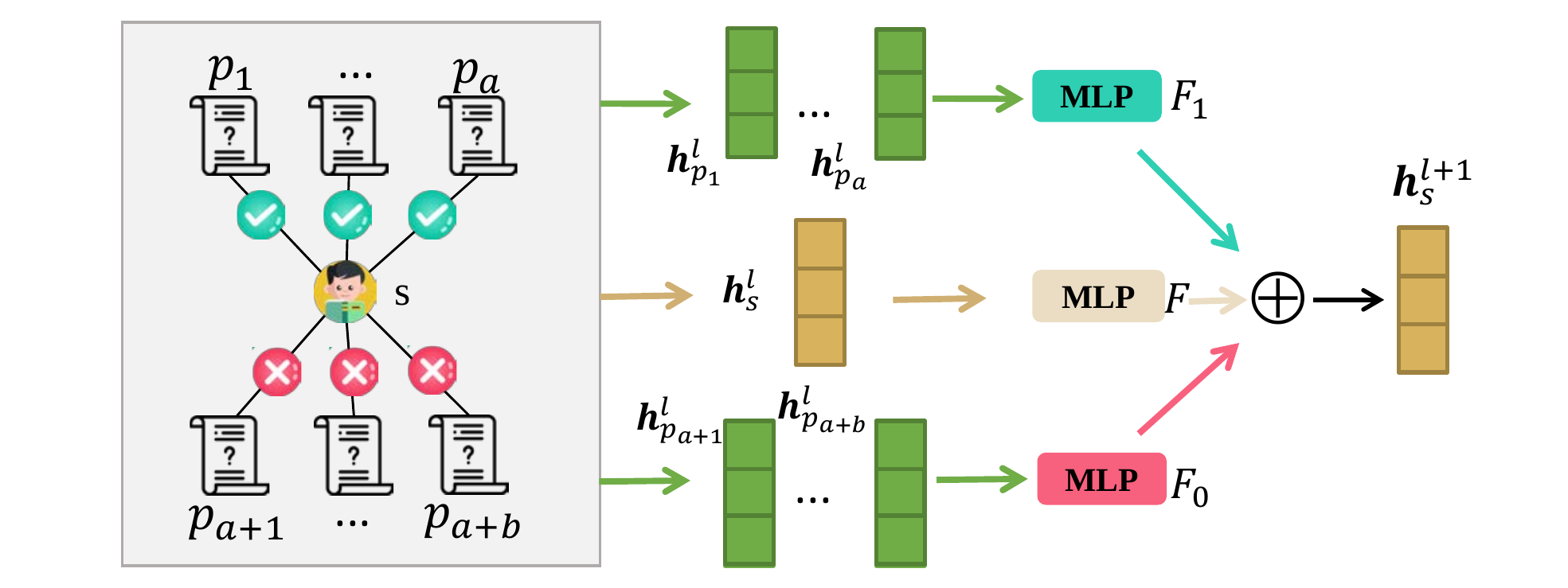}}
\caption{The graph convolution layer in our model. Suppose student $s$ has good performance scores on $a$ exercises and bad performance scores on other $b$ exercises. We show the information propagation process of the graph convolution layer on student $s$. } \label{fig2}
\end{figure}

Fig. \ref{fig2}. provides an overview of the one-layer GCN propagation process in our model. 
Specifically, we utilize link-specific aggregation functions based on multilayer perceptrons (MLPs) for two kinds of links, thus getting high-order students' and exercises' embeddings. Students' and exercises' initial embedding can be formulated as
\begin{equation}
\mathbf{h}_s^0 =\mathbf{u}_s, \mathbf{h}_p^0 =\mathbf{v}_p. 
\end{equation}
Suppose there are $L$ propagation layers. The student $s$'s embedding at the $(l+1)$-th layer can be formulated as: 
\begin{equation}
 \mathbf{h}_{s}^{(l+1)} = \sum_{n\in\{ 0,1\}}\sum_{p\in \mathcal{N}_s^n}\frac{1}{ |\mathcal{N}_s^n|}F_{n}(\mathbf{h}_{p}^{l})+F(\mathbf{h}_{s}^l),
 \label{eq2}
\end{equation}
where $\mathcal{N}_{s}^n$ denotes student $s$'s n-th type of neighbors, i.e.,the exercises that student $s$ answered correctly and incorrectly.
Eq.(\ref{eq2}) uses three types of functions $F,F_0,F_1$ to differentiate the aggregation process of student $s$'s neighbors and itself with MLPs:
\begin{equation}
\label{MLP_F}
F_n(\mathbf{h}^l) = \sigma_2(\sigma_1(\mathbf{h}^lW_1))W_2, 
\end{equation}
where $\sigma_1,\sigma_2$ denote activation functions and $W_1,W_2$ denote the linear transformation. The exercise $p$'s embedding can be obtained by aggregating embeddings of its neighbor nodes in the same way. 
After propagating information, we can obtain $[\mathbf{h}_s^0 ,...,\mathbf{h}_s^l,..,\mathbf{h}_s^L]$ and $[\mathbf{h}_p^0 ,...,\mathbf{h}_p^l,..,\mathbf{h}_p^L]$ as students' and exercises' embeddings of each layer. 
We concatenate embeddings of each layer as follows: 
\begin{equation}
\mathbf{h}_s = \mathbf{h}_s^0 ||... || \mathbf{h}_s^l||... ||\mathbf{h}_s^L,~~~\mathbf{h}_p = \mathbf{h}_p^0||... || \mathbf{h}_p^l||... || \mathbf{h}_p^L,
\end{equation}
where $||$ is the concatenation operation. By calculating the inner dot of concatenated embeddings, we can get student $s$'s proficiency on exercise $p$ in the exercise space: 
\begin{equation}
\hat{r}_{sp}^\mathcal{P} = \mathbf{h}_s^T \mathbf{h}_p.
\end{equation}

\subsection{Modeling Information of Knowledge Concepts}
\label{scorek}
As for predicting students' proficiency on knowledge concepts, we project students and knowledge concepts to knowledge-space. 
Note that, $\mathbf{X}=[\mathbf{x}_1,...,\mathbf{x}_s,...,$ $\mathbf{x}_M]\in \mathbb{R}^{M\times D}$ and  $\mathbf{Y}=[\mathbf{y}_1,...,\mathbf{y}_k,...,\mathbf{y}_O]\in \mathbb{R}^{O\times D}$ respectively denote the representations of students and knowledge concepts. 
Then, we use the inner product to predict $\hat{r}_{sp}^\mathcal{K}$ in the knowledge concept space:
\begin{equation}
\label{rspK}
 \hat{r}_{sp}^\mathcal{K}=\frac{1}{|\mathcal{K}_p|}\sum_{k\in \mathcal{K}_p} \mathbf{x}_s^\mathrm{T} \mathbf{y}_{k},
\end{equation}
where $|\mathcal{K}_p|$ denotes the set of knowledge concepts related to exercise $p$. It can be formulated as $\mathcal{K}_p=\{k | q_{pk}=1 \}$, while~$q_{pk} \in \mathbf{Q}$.
Please note that, we do not utilize GCN layer here. In fact, the number of knowledge concepts is much smaller than that of exercises. 
Thus, relations between students and knowledge concepts are not sparse enough. 
Therefore, we keep their original embeddings rather than utilizing GCN layers in Eq.(\ref{rspK}) to avoid over smoothing. 
\subsection{Performance Prediction}
In subsection \ref{sec3.3}, we model the effect of exercises by R-GCN technique. In subsection \ref{scorek},  we model the effect of knowledge concepts by inner product of student and knowledge concepts related to the target exercise.
After obtaining $\hat{r}_{sp}^\mathcal{P}$ and $\hat{r}_{sp}^\mathcal{K}$, we can easily calculate student $s$'s final predicted scores on exercise $p$. 
Thus the overall predicted function is defined as:
\begin{equation}
\label{EQ:RSP}
\hat{r}_{sp}= \hat{r}_{sp}^\mathcal{P}+\alpha \hat{r}_{sp}^\mathcal{K},
\end{equation}
where $\alpha$ is a hyper-parameter used to control the balance between the two sub-tasks. We use the point-wise based squared
loss to optimize our model:
\begin{equation}
\label{loss_function}
L = \frac{1}{T} \sum_{(s,p,r_{sp}) \in (S,P,\mathbf{R})}(r_{sp}-\hat{r}_{sp})^2, 
\end{equation}
where $T$ denotes the number of $(s,p,r_{sp})$ triplets in training data. 
\section{Experiments}
\subsection{Dataset Description}
\begin{table}[htb]
\caption{The statistics of the two datasets}\label{tab1}
\centering
\begin{tabular}{c|c|c|c|c|c}
\hline
Dataset & Students & Exercises &  Concepts & Logs & Density \\ \hline
ASISST & 4,163 & 17,746  & 123 & 278,868  &0.37\% \\ \hline
KDDcup & 574 & 173,650   & 437  & 609,979  & 0.61\%  \\ \hline
\end{tabular}
\end{table}
We choose two widely-used datasets in our experiments. One dataset is ASSIST~(ASSISTments 2009-2010 "skill builder")\footnote[1]{https://sites.google.com/site/assistmentsdata/home} provided by the online educational service ASSISTments. The other dataset is Algebra 2005-2006 from the Educational Data Mining Challenge of KDDCup\footnote[2]{http://pslcdatashop.web.cmu.edu/KDDCup/}. The detailed statistics of two datasets are summarized in Table~\ref{tab1}. Note that, we only consider exercises that are related to at least one knowledge concept. Specifically, we filter out exercises without related knowledge concepts for the two datasets.
Because the number of concepts in KDDcup dataset is too small to make full use of concept information, we combine the knowledge concepts related to the same exercise as a new single concept. 
Please note that, these two datasets are extremely different. Specifically, in ASISST, each student has nearly 67 logs on average while in KDDcup, each student has nearly 1,063 logs. 
\subsection{Experimental Settings}
\label{sub4.2}
\subsubsection{Evaluation Metrics.}
We adopt three widely used metrics~(Accuracy, RMSE, and AUC) to measure the error between true ratings and predicted ratings~\cite{wang2020neural,gkt,ekt}. 
Root mean square error~(RMSE) is used to measure the absolute difference between predicted labels and real labels~\cite{rmse}. As students' performance scores are binary, we utilize the area under the curve~(AUC)~\cite{auc} as a metric.
\subsubsection{Baselines.}
\label{sets}
We compare our model with the following methods:
\begin{itemize}
\item[$\bullet$] \textbf{Student Average}: This method 
calculates students' average scores in training data and uses them as the predicted scores on exercises in testing data. 

\item[$\bullet$] \textbf{MF}~\cite{mf}: 
It is a classical CF model in recommendations. This model utilizes MF techniques and learns latent representations of students and exercises. 
Note that knowledge concepts are not used in MF. 
\item[$\bullet$] \textbf{IRT}~\cite{irt}: A classical cognitive diagnosis model that uses one-dimensional continuous variables to represent students' knowledge proficiency and exercises' difficulty, and uses the difference between them for score prediction.
\item[$\bullet$] \textbf{NeuralCDM}~\cite{wang2020neural}: This is an improved multi-dimensional cognitive diagnosis model that utilizes neural networks as the interaction function.
\item[$\bullet$] \textbf{CRMF}~\cite{crmf}: This MF based model takes knowledge concepts into consideration by assuming that representations of exercises with the same knowledge concepts are more similar.
\item[$\bullet$] \textbf{R-GCN}~\cite{rgcn}: A substructure of our model that only uses R-GCN to predict scores but neglects students' proficiency on knowledge concepts. 
\item[$\bullet$] \textbf{R-GCN~(hetero)}: In this method, We apply GCN in the whole student-exercise-knowledge heterogeneous graph in Fig. \ref{fig:hetero}.
\end{itemize}
\subsubsection{Parameter Settings.} We implement our model in
PyTorch-1.6.0. The embedding dimension is set to 128 for our model and other CF models. We initialize all parameters with Xavier initialization~\cite{xavier}. The learning rate is set to 0.001. We set the depth of GCN as two layers. We choose 2-layer MLPs to serve as $F$ in Eq.(\ref{MLP_F}) and LeakyReLU as the activation function. We also set the balancing parameter $\alpha=1$. 
\subsection{Experimental Results} 
We list the results of our model and other baselines in Table \ref{tab:performance_comparison}. We have several observations from this table. 

First, Student Average is the simplest baseline. It assumes that students have the same scores on different exercises, resulting in the worst performance. Second, classical MF based models~(MF and CRMF) perform worse than our proposed \sname~on both two datasets. An obvious reason is that they ignore high-order collaborative information. Simultaneously, CRMF performs better than MF for considering course relations.
Third, as for two cognitive diagnosis models (NeuralCDM and IRT),
we observe that cognitive diagnosis models perform worse than all MF based model on the ASSIST dataset. 
MF based models explore similarity among students/exercises, and then provide suggested guidance for students based on the similarities. 
The reason is that 
lack of sufficient data brings trouble in cognitive diagnosis models while it has fewer effects on CF based models~(MF, CRMF, R-GCN, and \sname). 
Fourth, R-GCN~(hetero) even performs worse than R-GCN on KDDcup. The reason is that data in the heterogeneous graph doesn't obey power law distribution, therefore, common graph based methods cannot be directly applied onto ITSs  as mentioned in Section \ref{intro}.
Finally, our proposed \sname~has the best performance on both two datasets. 
The reason is that \sname~simultaneously utilizes information of high-order collaborative signals and  related knowledge concepts.

\begin{table}[]
  \centering  
  \caption{ Overall performance. $\uparrow/\downarrow$ denotes that the higher/lower, the better. }  
  \label{tab:performance_comparison}  
  \scalebox{0.9}{
  \begin{tabular}{|c|c|c|c|c|c|c|}
  \hline
  \multirow{2}{*}{Model} & \multicolumn{3}{c|}{ASSIST} & \multicolumn{3}{c|}{KDDcup} \\ \cline{2-7} 
   & ~~Accuracy~$\uparrow$~ & ~~RMSE~$\downarrow$~ & ~~AUC~$\uparrow$~ & ~~Accuracy~$\uparrow$~ & ~~RMSE~$\downarrow$~ & ~~AUC~$\uparrow$ ~\\ \hline
   Student Average&0.6942&0.4483&0.6816&0.7679&0.4190&0.5891 \\ \hline  
   MF&0.7399&0.4205&0.8105&0.7927&0.3841&0.8062\\ \hline  
   IRT&0.7181&0.4647&0.7394&0.7762&0.4835&0.7607\\ \hline  
   NeuralCDM&0.7249&0.4329&0.7561&0.8060&0.3713&0.8093\\ \hline  
    CRMF&0.7612&0.4134&0.8136&0.8014&0.3750&0.7968\\ \hline  
    R-GCN&0.7705&0.3982&0.8230&0.8205&0.3619&0.8239\\ \hline  
    R-GCN~(hetero)&0.7748&0.3973&0.8232&0.8201&0.3642&0.8187\\ \hline  
    \sname&{\bf 0.7782}&{\bf 0.3938}&{\bf 0.8298}&{\bf 0.8271}&{\bf 0.3591}&{\bf 0.8291}\\ \hline  
  \end{tabular}}
\end{table}

\subsection{Detailed Model Analyses}
\subsubsection{Ablation Study} We perform an ablation study to demonstrate the effectiveness of each component in our model. 
Specifically, we conduct four experiments to figure out whether graph based techniques~(denoted as GCN) and utilizing knowledge concepts~(KLG) are effective in ITSs. 
The basic model is MF, which only utilizes MF techniques.
Besides MF techniques, MF-TEM follows subsection \ref{scorek} to use knowledge concepts. 
MF and R-GCN are recorded in subsection \ref{sub4.2}. 
As shown in Table \ref{tab:Ablation}, MF-TEM performs better than MF. It proves that utilizing knowledge concepts is effective. Similarly, R-GCN also performs better than MF, which proves that capturing high-order collaborative signals is helpful for student score prediction. 
Finally, our proposed \sname~ performs best to prove that simultaneously considering R-GCN and knowledge concepts has the most performance improvements. 
  \label{tab:Ablation}  
  \begin{table}[]
  \centering  
  \caption{The ablation study}  
  \scalebox{0.85}{
  \begin{tabular}{|c|c|c|c|c|c|c|c|c|}
  \hline
  \multirow{2}{*}{Model} & \multicolumn{2}{c|}{Components} & \multicolumn{3}{c|}{ASSIST} & \multicolumn{3}{c|}{KDDcup} \\ \cline{2-9} 
   & KLG & GCN & ~Accuracy~$\uparrow$ & ~RMSE~$\downarrow$ & ~AUC~$\uparrow$ & ~Accuracy~$\uparrow$ & ~RMSE~$\downarrow$ & ~AUC~$\uparrow$ \\ \hline
   MF&$\times$&$\times$&0.7399&0.4205&0.8105&0.7927&0.3841&0.8062\\ \hline  
   MF-TEM&$\checkmark$&$\times$&0.7664&0.3984&0.8288&0.8255&0.3626&0.8246\\ \hline  
    R-GCN&$\times$&$\checkmark$&0.7705&0.3982&0.8230&0.8205&0.3619&0.8239\\ \hline  
    \sname&$\checkmark$&$\checkmark$&{\bf 0.7782}&{\bf 0.3938}&{\bf 0.8298}&{\bf 0.8271}&{\bf 0.3591}&{\bf 0.8291}\\ \hline  
  \end{tabular}}
\end{table}
\subsubsection{Performance under different balancing parameter $\alpha$} 
As mentioned in Eq. (\ref{EQ:RSP}), $\alpha$ is a hyper-parameter that controls balance between two-fold effects. We try the parameter $\alpha$ in the range \{0,0.1,1,5,10\}. Note that, $\alpha=0$ denotes only taking the exercise space into consideration, and \sname~ degenerates to R-GCN. 
As shown in Fig. \ref{fig:tradeoff}, when $\alpha \rightarrow 0$, the results become worse; when $\alpha$ becomes higher (e.g., $\alpha=5,10$), the results also become worse. Finally, \sname~ has the best performance on two datasets when $\alpha=1$.  
 \begin{small}
 \centering
 \begin{figure*} [htb]
 	    \subfigure[Performance on ASSIST]{
 	    \includegraphics[width=55mm]{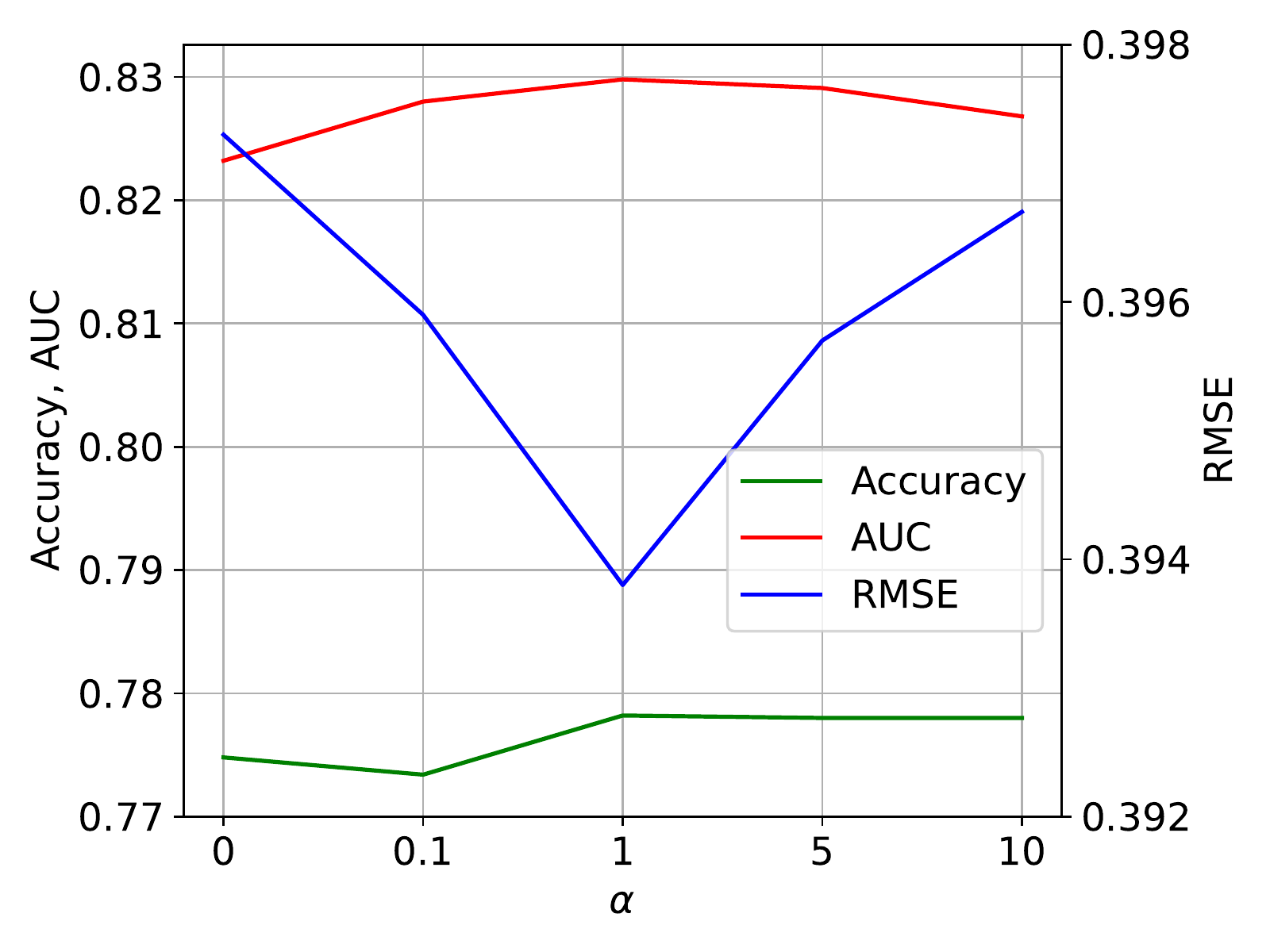}
 	    }
 	    \subfigure[Performance on KDDcup]{
 	    \includegraphics[width=55mm]{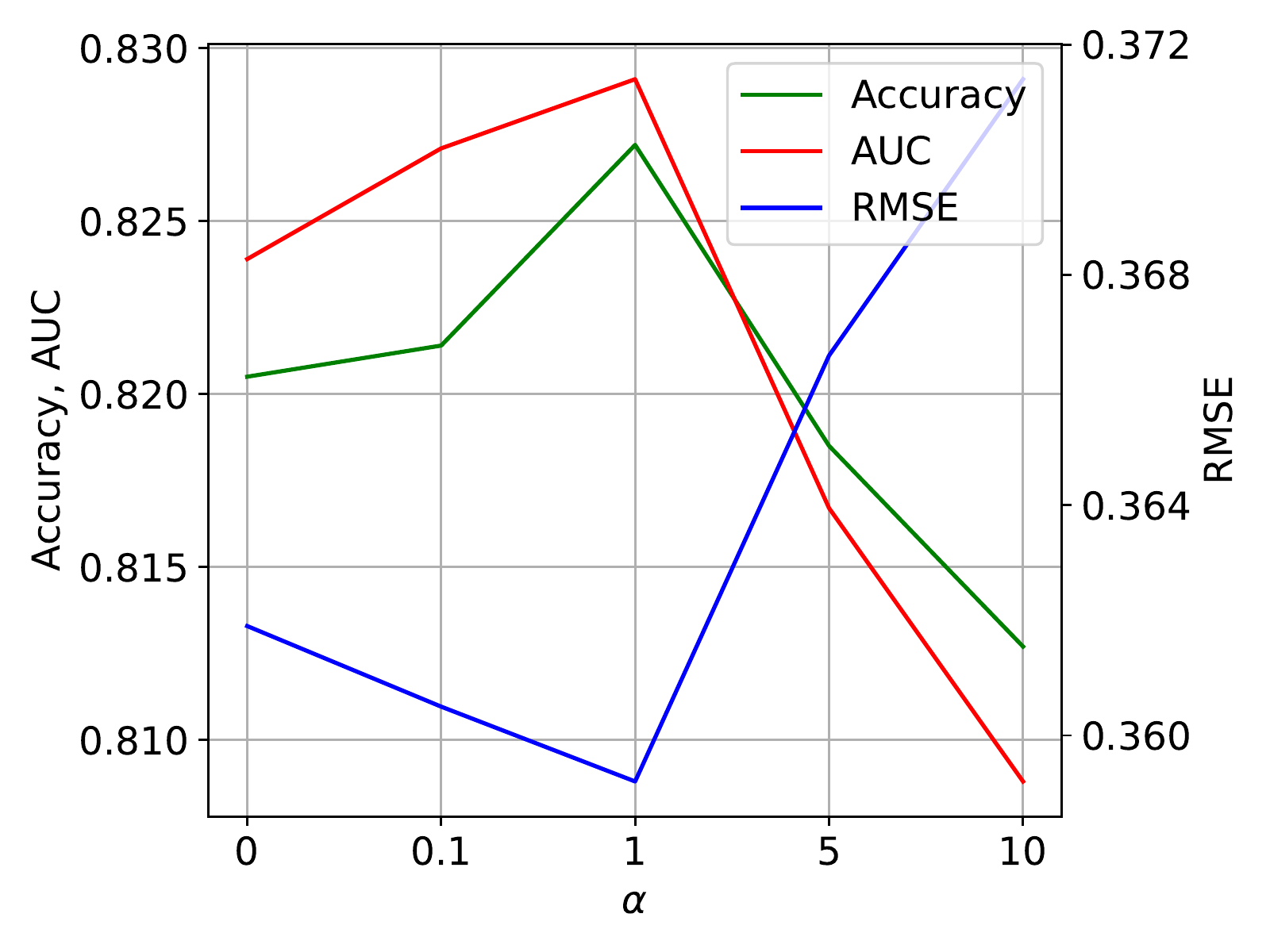}
 	    }
 		\caption{\small{Results of accuracy, AUC, and RMSE with different $\alpha$}} \label{fig:tradeoff}
 \end{figure*}
\end{small}
\section{Conclusions}
In this paper, we proposed a \fname~to improve student performance prediction in ITSs.
We borrowed the success of neural graph based models in recommender systems and successfully modeled two-fold effects of exercises and related knowledge concepts. 
Experimental results on two datasets showed the effectiveness of our model.
Note that, we assumed that students' knowledge states are static in this paper. In the future, we are interested in extending our model to a dynamic model which can track the changes in knowledge states.

\subsubsection{Acknowledgments.}This work is supported in part by by grants from iFLYTEK, P.R. China (Grant No. COGOS-20190002) and the Open Project Program of the National Laboratory of Pattern Recognition (NLPR).

\newpage
\bibliographystyle{splncs04}
\bibliography{mybib}

\begin{thebibliography}{10}
\providecommand{\url}[1]{\texttt{#1}}
\providecommand{\urlprefix}{URL }
\providecommand{\doi}[1]{https://doi.org/#1}

\bibitem{an2019neural}
An, M., Wu, F., Wu, C., Zhang, K., Liu, Z., Xie, X.: Neural news recommendation
  with long-and short-term user representations. In: Proceedings of the 57th
  Annual Meeting of the Association for Computational Linguistics. pp. 336--345
  (2019)

\bibitem{engaging}
Anderson, A., Huttenlocher, D., Kleinberg, J., Leskovec, J.: Engaging with
  massive online courses. In: Proceedings of WWW. pp. 687--698 (2014)

\bibitem{rmse}
Barnston, A.G.: Correspondence among the correlation, rmse, and heidke forecast
  verification measures; refinement of the heidke score. Weather and
  Forecasting  \textbf{7}(4),  699--709 (1992)

\bibitem{auc}
Bradley, A.P.: The use of the area under the roc curve in the evaluation of
  machine learning algorithms. Pattern recognition  \textbf{30}(7),  1145--1159
  (1997)

\bibitem{itsbook}
Burns, H., Luckhardt, C.A., Parlett, J.W., Redfield, C.L.: Intelligent tutoring
  systems: Evolutions in design. Psychology Press (2014)

\bibitem{lrgccf}
Chen, L., Wu, L., Hong, R., Zhang, K., Wang, M.: Revisiting graph based
  collaborative filtering: A linear residual graph convolutional network
  approach. In: Proceedings of AAAI. vol.~34, pp. 27--34 (2020)

\bibitem{chen2021set2setrank}
Chen, L., Wu, L., Zhang, K., Hong, R., Wang, M.: Set2setrank: Collaborative set
  to set ranking for implicit feedback based recommendation. arXiv preprint
  arXiv:2105.07377  (2021)

\bibitem{dirt}
Cheng, S., Liu, Q., Chen, E., Huang, Z., Huang, Z., Chen, Y., Ma, H., Hu, G.:
  Dirt: Deep learning enhanced item response theory for cognitive diagnosis.
  In: Proceedings of CIKM. pp. 2397--2400 (2019)

\bibitem{bkt}
Corbett, A.T., Anderson, J.R.: Knowledge tracing: Modeling the acquisition of
  procedural knowledge. User modeling and user-adapted interaction
  \textbf{4}(4),  253--278 (1994)

\bibitem{dina}
De~La~Torre, J.: Dina model and parameter estimation: A didactic. Journal of
  educational and behavioral statistics  \textbf{34}(1),  115--130 (2009)

\bibitem{irt}
Embretson, S.E., Reise, S.P.: Item response theory. Psychology Press (2013)

\bibitem{addressing}
Feng, M., Heffernan, N., Koedinger, K.: Addressing the assessment challenge
  with an online system that tutors as it assesses. User modeling and
  user-adapted interaction  \textbf{19}(3),  243--266 (2009)

\bibitem{xavier}
Glorot, X., Bengio, Y.: Understanding the difficulty of training deep
  feedforward neural networks. In: Proceedings of AISTATS. pp. 249--256 (2010)

\bibitem{learningorforgetting}
Huang, Z., Liu, Q., Chen, Y., Wu, L., Xiao, K., Chen, E., Ma, H., Hu, G.:
  Learning or forgetting? a dynamic approach for tracking the knowledge
  proficiency of students. ACM Transactions on Information Systems (TOIS)
  \textbf{38}(2),  1--33 (2020)

\bibitem{crmf}
Huynh-Ly, T.N., Le, H.T., Nguyen, T.N.: Integrating courses' relationship into
  predicting student performance. International Journal  \textbf{9}(4) (2020)

\bibitem{mf}
Koren, Y., Bell, R., Volinsky, C.: Matrix factorization techniques for
  recommender systems. Computer  \textbf{42}(8),  30--37 (2009)

\bibitem{ekt}
Liu, Q., Huang, Z., Yin, Y., Chen, E., Xiong, H., Su, Y., Hu, G.: Ekt:
  Exercise-aware knowledge tracing for student performance prediction. IEEE
  Transactions on Knowledge and Data Engineering  \textbf{33}(1),  100--115
  (2019)

\bibitem{fuzzy}
Liu, Q., Wu, R., Chen, E., Xu, G., Su, Y., Chen, Z., Hu, G.: Fuzzy cognitive
  diagnosis for modelling examinee performance. ACM Transactions on Intelligent
  Systems and Technology  \textbf{9}(4),  1--26 (2018)

\bibitem{gkt}
Nakagawa, H., Iwasawa, Y., Matsuo, Y.: Graph-based knowledge tracing: modeling
  student proficiency using graph neural network. In: IEEE/WIC/ACM
  International Conference on Web Intelligence. pp. 156--163 (2019)

\bibitem{rkt}
Pandey, S., Srivastava, J.: Rkt: Relation-aware self-attention for knowledge
  tracing. In: Proceedings of CIKM. pp. 1205--1214 (2020)

\bibitem{dkt}
Piech, C., Spencer, J., Huang, J., Ganguli, S., Sahami, M., Guibas, L.,
  Sohl-Dickstein, J.: Deep knowledge tracing. arXiv preprint arXiv:1506.05908
  (2015)

\bibitem{edm}
Romero, C., Ventura, S., Pechenizkiy, M., Baker, R.S.: Handbook of educational
  data mining. CRC press (2010)

\bibitem{rgcn}
Schlichtkrull, M., Kipf, T.N., Bloem, P., Van Den~Berg, R., Titov, I., Welling,
  M.: Modeling relational data with graph convolutional networks. In: ESWC. pp.
  593--607 (2018)

\bibitem{convolutionalkt}
Shen, S., Liu, Q., Chen, E., Wu, H., Huang, Z., Zhao, W., Su, Y., Ma, H., Wang,
  S.: Convolutional knowledge tracing: Modeling individualization in student
  learning process. In: Proceedings of the 43rd International ACM SIGIR
  Conference on Research and Development in Information Retrieval. pp.
  1857--1860 (2020)

\bibitem{rspsp}
Thai-Nghe, N., Drumond, L., Krohn-Grimberghe, A., Schmidt-Thieme, L.:
  Recommender system for predicting student performance. Procedia Computer
  Science  \textbf{1}(2),  2811--2819 (2010)

\bibitem{multi}
Thai-Nghe, N., Schmidt-Thieme, L.: Multi-relational factorization models for
  student modeling in intelligent tutoring systems. In: IEEE KSE. pp. 61--66
  (2015)

\bibitem{social}
Thanh-Nhan, H.L., Huy-Thap, L., Thai-Nghe, N.: Toward integrating social
  networks into intelligent tutoring systems. In: IEEE KSE. pp. 112--117 (2017)

\bibitem{wang2020neural}
Wang, F., Liu, Q., Chen, E., Huang, Z., Chen, Y., Yin, Y., Huang, Z., Wang, S.:
  Neural cognitive diagnosis for intelligent education systems. In: Proceedings
  of AAAI. vol.~34, pp. 6153--6161 (2020)

\bibitem{ngcf}
Wang, X., He, X., Wang, M., Feng, F., Chua, T.S.: Neural graph collaborative
  filtering. In: Proceedings of ACM SIGIR. pp. 165--174 (2019)

\bibitem{wu2021survey}
Wu, L., He, X., Wang, X., Zhang, K., Wang, M.: A survey on neural
  recommendation: From collaborative filtering to content and context enriched
  recommendation. arXiv preprint arXiv:2104.13030  (2021)

\bibitem{wu2020diffnet++}
Wu, L., Li, J., Sun, P., Hong, R., Ge, Y., Wang, M.: Diffnet++: A neural
  influence and interest diffusion network for social recommendation. IEEE
  Transactions on Knowledge and Data Engineering  (2020)

\bibitem{diffnet}
Wu, L., Sun, P., Fu, Y., Hong, R., Wang, X., Wang, M.: A neural influence
  diffusion model for social recommendation. In: Proceedings of the 42nd
  international ACM SIGIR conference on research and development in information
  retrieval. pp. 235--244 (2019)

\bibitem{wu2020learning}
Wu, L., Yang, Y., Chen, L., Lian, D., Hong, R., Wang, M.: Learning to transfer
  graph embeddings for inductive graph based recommendation. In: Proceedings of
  the 43rd International ACM SIGIR Conference on Research and Development in
  Information Retrieval. pp. 1211--1220 (2020)

\bibitem{wu2020joint}
Wu, L., Yang, Y., Zhang, K., Hong, R., Fu, Y., Wang, M.: Joint item
  recommendation and attribute inference: An adaptive graph convolutional
  network approach. In: Proceedings of the 43rd International ACM SIGIR
  Conference on Research and Development in Information Retrieval. pp. 679--688
  (2020)

\bibitem{Mt-mcd}
Zhu, T., Liu, Q., Huang, Z., Chen, E., Lian, D., Su, Y., Hu, G.: Mt-mcd: A
  multi-task cognitive diagnosis framework for student assessment. In:
  International Conference on Database Systems for Advanced Applications. pp.
  318--335. Springer (2018)

\end{thebibliography}





\end{document}